\documentclass[11pt]{article}

\usepackage[final]{acl}

\usepackage{times}
\usepackage{latexsym}

\usepackage[T1]{fontenc}

\usepackage[utf8]{inputenc}

\usepackage{fontawesome5}

\usepackage{microtype}

\usepackage{inconsolata}

\usepackage{graphicx}

\usepackage{amsmath}
\usepackage{xspace}
\usepackage{amsfonts}
\usepackage{amssymb}
\usepackage{algorithm}
\usepackage{algpseudocode}
\usepackage[most]{tcolorbox}
\usepackage{tabularx}

\usepackage{booktabs}
\usepackage{multirow}
\usepackage{xcolor}
\usepackage{colortbl}
\usepackage{array}
\usepackage{makecell}
\usepackage{siunitx}
\usepackage{subcaption}
\usepackage{pifont}
\usepackage{fontawesome5}
\usepackage{enumitem}
\usepackage{bbm}
\usepackage{listings}

\usepackage{amsthm}
\usepackage{bm}

\definecolor{sedimablue}{HTML}{315A7D}
\definecolor{sedimalight}{HTML}{F3F7FA}
\definecolor{sedimaborder}{HTML}{A8BCCB}
\definecolor{codebackground}{HTML}{F7F8FA}
\definecolor{codekeyword}{HTML}{7A3E9D}
\definecolor{codecomment}{HTML}{5F7F67}
\definecolor{codestring}{HTML}{A34A28}

\lstdefinestyle{sedimapy}{
    language=Python,
    basicstyle=\ttfamily\fontsize{7.4}{8.7}\selectfont,
    keywordstyle=\color{codekeyword}\bfseries,
    commentstyle=\color{codecomment},
    stringstyle=\color{codestring},
    showstringspaces=false,
    columns=fullflexible,
    keepspaces=true,
    tabsize=4,
    breaklines=true,
    breakatwhitespace=false,
    postbreak=\mbox{\textcolor{gray}{$\hookrightarrow$}\space},
    numbers=left,
    numberstyle=\ttfamily\tiny\color{gray},
    numbersep=9pt,
    xleftmargin=3.2em,
    framexleftmargin=0pt,
    resetmargins=true,
    aboveskip=0pt,
    belowskip=0pt
}

\newtcblisting{sedimacode}[2][]{
    enhanced,
    listing only,
    listing engine=listings,
    listing options={
        style=sedimapy,
        #1
    },
    colback=codebackground,
    colframe=sedimaborder,
    boxrule=0.7pt,
    arc=2mm,
    outer arc=2mm,
    boxsep=0mm,
    left=1.5mm,
    right=1.5mm,
    top=1.5mm,
    bottom=1.5mm,
    title={#2},
    fonttitle=\small\bfseries,
    coltitle=white,
    colbacktitle=sedimablue,
    titlerule=0pt,
    halign title=left,
    lefttitle=2mm,
    righttitle=2mm,
    toptitle=1.2mm,
    bottomtitle=1.2mm,
    before skip=8pt,
    after skip=8pt
}

\newtcolorbox{sedimabox}[2][]{
    enhanced,
    breakable,
    colback=sedimalight,
    colframe=sedimaborder,
    boxrule=0.7pt,
    arc=2mm,
    left=3mm,
    right=3mm,
    top=2mm,
    bottom=2mm,
    title={#2},
    fonttitle=\small\bfseries,
    coltitle=white,
    colbacktitle=sedimablue,
    attach boxed title to top left={xshift=2mm,yshift=-2mm},
    boxed title style={arc=1.5mm,boxrule=0pt},
    #1
}

\newcounter{prompt}[section]
\renewcommand{\theprompt}{\thesection.\arabic{prompt}}

\newenvironment{promptbox*}[2][]{%
    \begin{figure*}[t]
    \refstepcounter{prompt}%
    \begin{tcolorbox}[
        colback=blue!5!white,
        colframe=blue!75!black,
        fonttitle=\bfseries,
        title={Prompt \theprompt: #2},
        enhanced,
        width=\textwidth,
        attach boxed title to top left={
            yshift=-2mm,
            xshift=5mm
        },
        boxed title style={
            colback=blue!75!black
        },
        sharp corners=south,
        drop shadow,
        left=1mm,
        right=1mm,
        top=2mm,
        bottom=1mm,
        boxsep=1mm,
        #1
    ]%
    \ttfamily\small
    \linespread{1.2}\selectfont
    \obeylines
    \obeyspaces
    \ignorespaces
}{%
    \end{tcolorbox}
    \end{figure*}
}

\newcommand{\sysname}{\textsc{Sedima}}

\title{\sysname: Cross-Run Hierarchical Insight Memory for\\ Evolutionary Search Agents}

\author{
\textbf{
Amirhossein Abaskohi$^{1}$\thanks{Corresponding author: \texttt{aabaskoh@cs.ubc.ca}.},
Mahdi Mostajabdaveh$^{2}$,
Zirui Zhou$^{2}$}\\[0.5em]
$^{1}$University of British Columbia, $^{2}$Huawei Technologies Canada
}
\begin{document}
\maketitle

\begin{abstract}
Large language model (LLM)-driven evolutionary search is a powerful paradigm for automated program and algorithm discovery, yet existing systems are largely \emph{memoryless}: each run explores from scratch, so agents repeatedly rediscover the same improvements and re-encounter the same dead ends. We introduce \sysname{}, a persistent hierarchical insight memory for evolutionary search agents. \sysname{} distills raw traces into natural-language insights, clusters them by semantic similarity using attention-weighted centroids, and retrieves relevant guidance to condition future mutations, accumulating transferable knowledge \emph{across} runs and problems rather than within a single trajectory. As a drop-in module that leaves the search operators unmodified, \sysname{} improves average final performance by \textbf{5.5\% on AlgoTune} and \textbf{6.6\% on ALE-Bench LITE} under a fixed budget of 100 evaluated candidates. Under OpenEvolve, \sysname{} reaches baseline-best performance with \textbf{32.3\% fewer iterations} on average across the five evaluated backbones\footnote{Code is available on \href{https://github.com/AmirAbaskohi/SEDIMA}{{\textcolor{black}{\faGithub}}~GitHub}.}.
\end{abstract}


\begin{figure*}
    \centering
    \includegraphics[width=\linewidth]{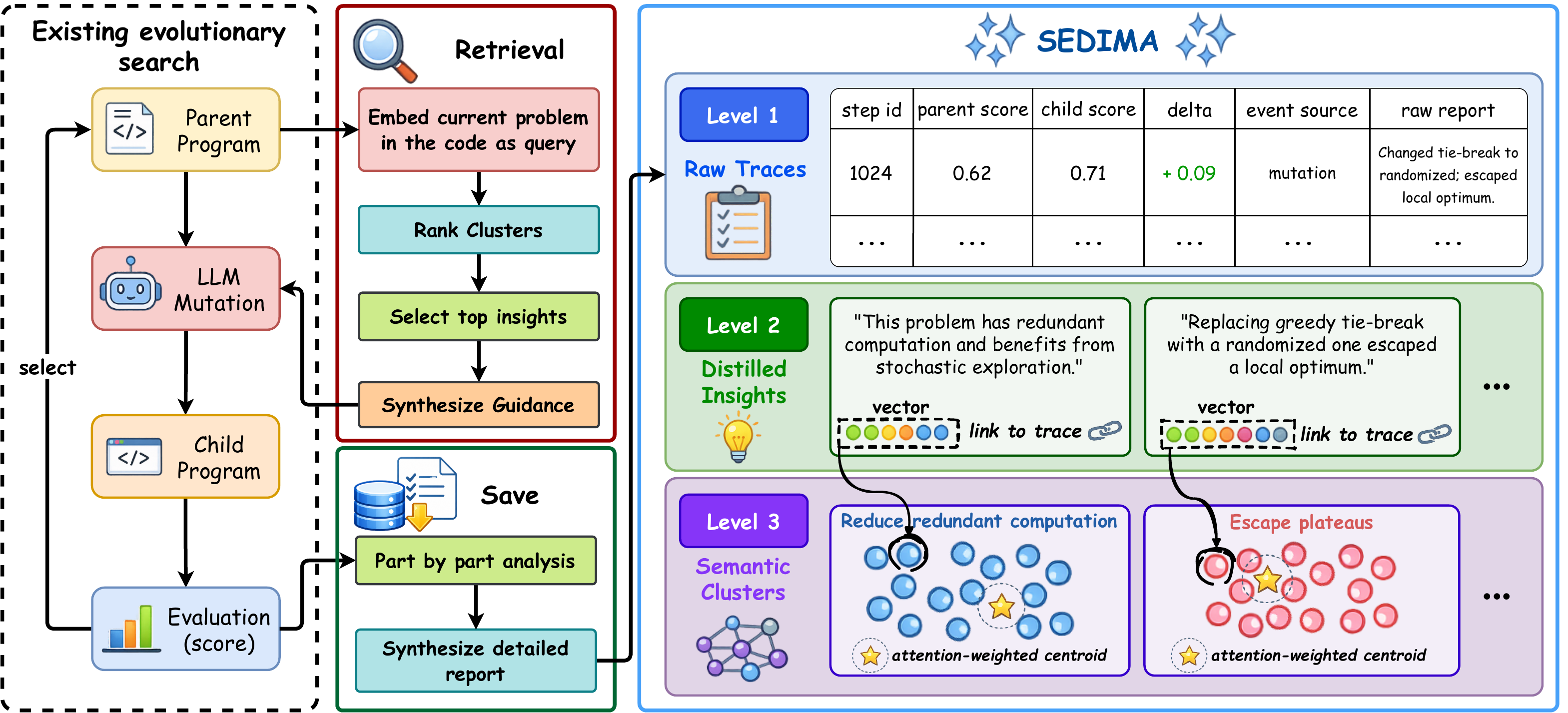}
    \caption{\textbf{Overview of \sysname{} .} A persistent three-level memory (right) augments an evolutionary search loop (left, dashed): evaluated children are distilled into hierarchical insights on \emph{save}, and relevant insights are injected into the mutation prompt on \emph{retrieval}.}
    \label{fig:method}
\end{figure*}

\section{Introduction}

Recent advances have turned LLMs from one-shot code generators into iterative search operators. Rather than producing a single solution, these systems repeatedly select promising candidates, use an LLM to propose semantically informed modifications, evaluate the resulting programs, and retain the strongest variants for further refinement \citep{FunSearch2023, novikov2025alphaevolvecodingagentscientific}. This formulation has enabled progress in mathematical discovery, algorithm design, and heuristic synthesis, while open-source frameworks such as OpenEvolve, ShinkaEvolve, and LLM4AD have made the approach increasingly accessible \citep{openevolve, lange2025shinkaevolveopenendedsampleefficientprogram, liu2026llm4adplatformalgorithmdesign}.

Despite this success, these systems are fundamentally \emph{memoryless}: each run begins from scratch and discards the reasoning that produced its successes and failures the moment the run terminates. This is wasteful in two ways. Across runs, an agent applied to a new task re-derives modifications it has already discovered and re-explores directions that were previously neutral or useless, because it retains no record of \emph{why} a past edit helped or hurt. Within a single run, the search trajectory is unstable: solution quality fluctuates sharply across iterations as the agent proposes, forgets, and re-proposes locally useful edits, with no persistent signal consolidating what is working. Recent work begins to close this gap by conditioning mutation on reflective signals rather than raw code \citep{ye2024reevo, LoongFlow2025, jiang2026deltaevolve}, but in each case memory remains within a single evolutionary trajectory and resets when a new run begins.

We introduce \sysname{}, a \textbf{persistent hierarchical insight memory for evolutionary search agents}. \sysname{} distills raw evolution traces into compact natural-language insights capturing why an edit succeeded or failed, then organizes them into a \textbf{three-level hierarchy}: raw traces, distilled insights, and semantically clustered insights whose cluster representatives are formed by \textbf{attention-weighted centroids}, so that prototypical, high-value lessons dominate retrieval. To condition a mutation, \sysname{} retrieves relevant guidance by semantic similarity and injects it into the prompt.

Two properties distinguish how this memory is used. First, it \textbf{persists across runs and problems}, so each run is warm-started with distilled experience rather than starting cold. Second, because insights are \textbf{retrieved by semantic content rather than genealogical lineage}, guidance learned on one problem can surface for another that is only loosely related, letting \textbf{experience transfer beyond a single task family}. Crucially, \sysname{} interfaces with the underlying search at exactly two points, a write on evaluation and a read into the mutation prompt, leaving selection, population management, and mutation operators unmodified. This makes it a \textbf{drop-in module} rather than a redesign: we integrate the same memory, unchanged, into two independent open-source harnesses, OpenEvolve~\cite{openevolve} and ShinkaEvolve~\cite{lange2025shinkaevolveopenendedsampleefficientprogram}, with no modification to either codebase's search logic.

\section{Related Work}
Prior work has begun to give evolutionary search agents a memory, but it stays confined to a single run or requires rearchitecting the search. ReEvo \citep{ye2024reevo} reflects over performance differences, LoongFlow \citep{LoongFlow2025} summarizes each generation into insights retrieved along the genealogical lineage, and DeltaEvolve \citep{jiang2026deltaevolve} stores pairwise semantic deltas; in all three, memory is scoped to one trajectory and discarded when the run ends, whereas \sysname{} persists insight across runs and problems and retrieves it by semantic similarity rather than lineage. CORAL \citep{qu2026coral} and PRISM \citep{mishra2026prismevolutionarymemorysubstrate} add persistent memory but only by turning the loop into long-running multi-agent systems, whereas \sysname{} is a single-agent, drop-in module that leaves the search operators unmodified. Beyond evolutionary search, general-purpose agents accumulate reusable knowledge across episodes as distilled insights \citep{Zhao_Huang_Xu_Lin_Liu_Huang_2024, ouyang2026reasoningbankscalingagentselfevolving}, skills \citep{wang2024voyager}, or verified experience libraries \citep{kong2025alphaopt}, and others meta-evolve the memory architecture \citep{zhang2026memevolve} or train a memory policy with dense rewards \citep{li2026sesearchselfevolvingsearchagent}; \sysname{} instead brings this distill-embed-retrieve recipe to population-based evolution with a non-parametric, hierarchically clustered store whose attention-weighted centroids \citep{goncalves-etal-2023-supervising} let high-value insights dominate retrieval. To our knowledge, \sysname{} is the first hierarchical memory for single-agent evolutionary search that transfers across runs and problems without changing search operators.

\section{\sysname}
\label{sec:method}

Figure~\ref{fig:method} shows how \sysname{} connects to evolutionary search through two hooks: a write after each evaluation and a read before each mutation. It leaves the search logic unchanged while maintaining a persistent three-level memory hierarchy that carries experience across runs and problems.

\noindent
\textbf{Level 1: raw traces.}
After each evaluation, \sysname{} stores a trace containing the step index, parent and child fitness, fitness difference, event source, and raw evaluation report. Level~1 performs no interpretation and serves as evidence for higher-level memories. Retaining these traces allows retrieved recommendations to remain linked to the evaluation events from which they were derived.

\noindent
\textbf{Level 2: distilled insights.}
An LLM converts each trace into a compact natural-language insight and links it back to its Level~1 evidence. For the initial evaluation of a problem, it generates a \emph{problem-level} insight describing the main algorithmic or architectural bottleneck. For each subsequent parent--child evaluation, it generates a \emph{change-level} insight describing the type of modification and its observed performance effect using the parent fitness, child fitness, fitness delta, change summary, and evaluation report. These abstractions remove implementation-specific details while preserving reusable lessons such as beneficial caching, pruning, vectorization, data-structure changes, or reductions in redundant computation. Each insight is embedded for semantic organization and retrieval.

\noindent
\textbf{Level 3: semantic clusters.}
Level~2 insights are grouped by embedding similarity: a new insight joins the nearest cluster if its cosine similarity exceeds a threshold and otherwise starts a new one. Cluster representatives are refreshed as members are added, and an optional LLM summary captures the recurring optimization pattern. Rather than a plain mean, \sysname{} uses a non-parametric attention-weighted centroid that assigns larger weights to semantically central, prototypical insights; lower temperatures sharpen this emphasis, while higher temperatures approach mean pooling. Appendix~\ref{sec:memory_details} provides the formal definitions.

\noindent
\textbf{Hierarchical retrieval.}
Before mutation, \sysname{} embeds a query describing the current program and its observed issues, retrieves the $K_c$ most similar clusters above a threshold, and then selects the $K_i$ most relevant insights from each. Retrieved items retain cluster- and insight-level similarity scores and may include up to two linked Level~1 traces. An LLM converts this evidence into 3 concrete recommendations inserted into the diff or full-rewrite prompt without changing the harness logic. Because retrieval is semantic rather than genealogical, lessons can transfer across problems with similar failure modes or optimization patterns.

\noindent
\textbf{Budgeting and fallback.}
Retrieval is bounded by fixed limits on clusters, insights, and prompt length, so context cost remains constant as memory grows. If no cluster passes the threshold, \sysname{} falls back to the original mutation prompt. Each write adds one distillation call and one insight embedding, while each memory-conditioned mutation adds one query embedding and one recommendation-synthesis call; optional cluster summaries require only periodic write-time calls. Appendix~\ref{sec:memory_details} provides the full procedure.

\section{Experiments}
\label{sec:experiments}

\subsection{Experimental Setup}
\label{sec:setup}

\paragraph{Benchmarks and evaluation protocol.}
We evaluate \sysname{} on AlgoTune \citep{NEURIPS2025_9a648e8e}, reporting the harmonic mean speedup over numerical programming tasks, and the ten-problem ALE-Bench LITE subset \citep{imajuku2025alebench}, reporting its problem-agnostic average performance score and separate short- and long-format results. Search uses only public local-evaluation scores as fitness, while all reported results come from private evaluation. We integrate \sysname{} into OpenEvolve \citep{openevolve} and ShinkaEvolve \citep{lange2025shinkaevolveopenendedsampleefficientprogram} without changing their selection, population, evaluation, or mutation logic. Each run is limited to 100 evolution steps with one evaluated child per step. Comparisons are therefore matched by the number of evaluated candidates rather than total LLM calls or tokens; \sysname{} incurs additional memory-construction and recommendation-synthesis calls as described in Section~\ref{sec:method}. Main-table cells use one run, while the per-step, ablation, and transfer experiments use three seeds. Memory is populated using disjoint problems, ensuring that no evaluation problem contributes to the memory used to solve it.

\paragraph{Models and implementation.}
We evaluate GPT-5.4 \citep{openai2026gpt54}, DeepSeek V4 Pro \citep{deepseekai2026deepseekv4highlyefficientmilliontoken}, Gemini 3 Pro \citep{google2025gemini3pro}, Qwen 3.7 Max \citep{qwen2026qwen37max}, and Qwen 3 Coder \citep{cao2026qwen3codernexttechnicalreport}. All LLM calls within a run use the same backbone, preventing stronger models from being used for memory construction or retrieval. We use a temperature of $0.7$ for mutation and $0.2$ for insight distillation, cluster summarization, and recommendation synthesis. Embeddings are produced by Qwen-Embedding-4B~\citep{zhang2025qwen3embeddingadvancingtext} and fixed across conditions. We set the cluster threshold to $\tau_{\mathrm{cluster}}=0.8$, retrieval threshold to $\tau_{\mathrm{ret}}=0.6$, and centroid temperature to $\tau=0.1$, and retrieve up to $K_c=3$ clusters with $K_i=3$ insights per cluster. Retrieved evidence is limited to 2,500 tokens across all models and harnesses. Each insight may include up to two linked traces, and the synthesizer returns three recommendations. Appendix~\ref{sec:setup_details} provides the full configuration. See Appendix~\ref{app:prompts} for the prompts used for \sysname{}.

\subsection{Effectiveness Across Models and Search Harnesses}
\label{sec:main_results}

Tables~\ref{tab:algotune} and~\ref{tab:alebench} show that \sysname{} improves every model under both OpenEvolve and ShinkaEvolve. Across the 20 matched model--harness--benchmark comparisons, all 20 favor \sysname{} (two-sided exact sign test, $p=1.9\times10^{-6}$), showing a consistent direction of improvement across configurations, although individual main-table cells remain single-run results. Averaged across harnesses, \sysname{} raises performance by \textbf{5.5\%} on AlgoTune and \textbf{6.6\%} on ALE-Bench LITE, despite ShinkaEvolve generally providing stronger baselines. The largest relative gains occur for Qwen 3 Coder, whose ALE-Bench score improves by 11.0\% under OpenEvolve and 9.0\% under ShinkaEvolve, suggesting that persistent guidance is especially valuable when the underlying search is weaker. The improvements also hold for both short- and long-format ALE-Bench problems, with larger gains on short contests where recurring optimization patterns are more transferable; the format-level breakdown and per-problem results are reported in Appendices~\ref{sec:alebench_format} and~\ref{sec:per_problem_performance}.

\begin{table}[t]
    \centering
    \footnotesize
    \caption{AlgoTune, harmonic mean speedup.}
    \label{tab:algotune}
    \resizebox{\linewidth}{!}{%
    \begin{tabular}{lcccc}
        \toprule
        & \multicolumn{2}{c}{OpenEvolve} & \multicolumn{2}{c}{ShinkaEvolve} \\
        \cmidrule(lr){2-3} \cmidrule(lr){4-5}
        Model & Base & +\sysname{} & Base & +\sysname{} \\
        \midrule
        GPT-5.4 & 1.48 & 1.56 & 1.53 & 1.60 \\
        DeepSeek V4 Pro & 1.75 & 1.85 & 1.80 & 1.89 \\
        Gemini 3 Pro & 1.83 & 1.91 & 1.87 & 1.94 \\
        Qwen 3.7 Max & 1.69 & 1.80 & 1.74 & 1.84 \\
        Qwen 3 Coder & 1.44 & 1.55 & 1.49 & 1.59 \\
        \midrule
        Mean & 1.64 & 1.73 & 1.69 & 1.77 \\
        \bottomrule
    \end{tabular}
    }
\end{table}

\begin{table}[t]
    \centering
    \caption{ALE-Bench LITE, average performance.}
    \label{tab:alebench}
    \resizebox{\linewidth}{!}{%
    \begin{tabular}{lcccc}
        \toprule
        & \multicolumn{2}{c}{OpenEvolve} & \multicolumn{2}{c}{ShinkaEvolve} \\
        \cmidrule(lr){2-3} \cmidrule(lr){4-5}
        Model & Base & +\sysname{} & Base & +\sysname{} \\
        \midrule
        GPT-5.4 & 1613.1 & 1709.9 & 1660.0 & 1743.7 \\
        DeepSeek V4 Pro & 1007.2 & 1087.8 & 1055.1 & 1123.6 \\
        Gemini 3 Pro & 1182.8 & 1259.6 & 1215.3 & 1281.8 \\
        Qwen 3.7 Max & 1186.0 & 1275.0 & 1225.2 & 1304.6 \\
        Qwen 3 Coder & 482.5 & 535.6 & 510.0 & 555.9 \\
        \midrule
        Mean & 1094.3 & 1173.6 & 1133.1 & 1201.9 \\
        \bottomrule
    \end{tabular}
    }
\end{table}

\subsection{Search Dynamics}
\label{sec:dynamics}

\begin{figure}[t]
    \centering
    \includegraphics[width=\linewidth]{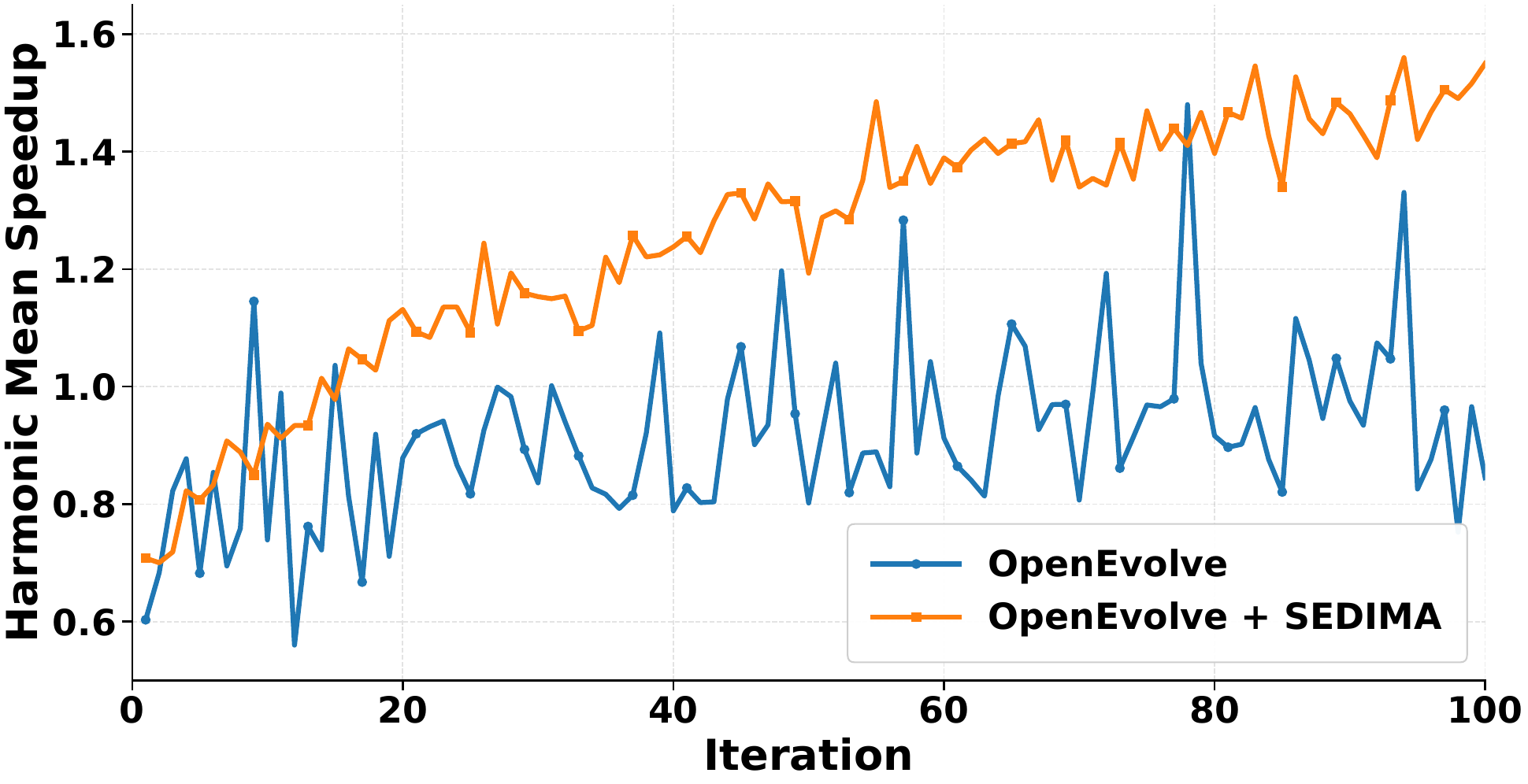}
    \caption{Mean AlgoTune speedup over 100 OpenEvolve iterations with GPT-5.4, with and without \sysname{}, averaged over three seeds.}
    \label{fig:per_step}
\end{figure}

Figure~\ref{fig:per_step} tracks AlgoTune score against evolution step over the full 100-step budget, isolating how quickly each configuration reaches a given quality level rather than only where it ends. With GPT-5.4 under OpenEvolve, \sysname{} first surpasses the baseline-best speedup of $1.48\times$ at step 55, whereas the baseline reaches this value at step 78. Thus, \sysname{} requires 23 fewer steps, corresponding to a \textbf{29.5\% reduction} in the iterations needed to reach baseline-best performance. The same trend holds across the remaining models: DeepSeek V4 Pro reaches its baseline-best level in 50 rather than 76 steps (\textbf{34.2\% fewer}), Gemini 3 Pro in 54 rather than 71 steps (\textbf{23.9\% fewer}), Qwen 3.7 Max in 52 rather than 84 steps (\textbf{38.1\% fewer}), and Qwen 3 Coder in 58 rather than 90 steps (\textbf{35.6\% fewer}). Averaged across the five backbones under OpenEvolve, \sysname{} requires \textbf{32.3\% fewer iterations} to reach baseline-best performance. These results indicate that persistent retrieval improves not only final solution quality, but also the rate at which evolutionary search discovers high-performing candidates; because the comparison is made by evolution step, this result should not be interpreted as an equivalent reduction in total LLM compute.

\subsection{Cross-Run Persistence and Transfer}
\label{sec:persistence}

Table~\ref{tab:transfer} varies memory provenance with evaluation problems fixed. In the cold condition, the memory is empty at deployment but continues to write and retrieve experience online within the run; comparing cold and warm memory therefore isolates the benefit of persistent experience from previous runs beyond within-run memory accumulation. Same-benchmark memory performs best, improving over the cold setting by \textbf{4.8\%} on AlgoTune and \textbf{6.7\%} on ALE-Bench. Memory collected from the other benchmark still yields gains of \textbf{1.8\%} and \textbf{2.9\%}, respectively, showing that \sysname{} transfers useful optimization knowledge across problem families, although in-domain experience remains more effective.

\begin{table}[t]
    \centering
    \footnotesize
    \caption{Effect of memory provenance at deployment.}
    \label{tab:transfer}
    \begin{tabular}{lcc}
        \toprule
        Memory at deployment & AlgoTune & ALE-Bench \\
        \midrule
        Cold (empty)          & 1.67 & 1113.7 \\
        Warm, same benchmark  & 1.75 & 1187.8 \\
        Warm, other benchmark & 1.70 & 1146.5 \\
        \bottomrule
    \end{tabular}
\end{table}

\paragraph{Additional Analysis}
A random-retrieval control performs only marginally above no memory (1.62 vs.\ 1.61 on AlgoTune-20 and 1102.6 vs.\ 1094.3 on ALE-Bench LITE), while full \sysname{} reaches 1.69 and 1173.6, respectively, indicating that the improvement primarily comes from retrieving relevant experience rather than simply adding memory text to the mutation prompt. Appendix~\ref{sec:ablations} additionally reports component and embedding-model ablations, while Appendix~\ref{sec:example} qualitatively traces an evolved program together with its corresponding memory write and retrieval.

\section{Conclusion}
\label{sec:conclusion}

We introduced \sysname{}, a persistent hierarchical insight memory for evolutionary search agents. \sysname{} distills raw traces into natural-language insights, organizes them using attention-weighted semantic clusters, and retrieves relevant guidance for future mutations. Through only two interfaces---a write after evaluation and a read before mutation---the same module integrates unchanged with OpenEvolve and ShinkaEvolve without modifying their search operators. Across both harnesses, \sysname{} improves average final performance by \textbf{5.5\% on AlgoTune} and \textbf{6.6\% on ALE-Bench LITE} under a fixed budget of 100 evaluated candidates. Under OpenEvolve, it requires \textbf{32.3\% fewer iterations} on average to reach baseline-best performance across the five evaluated backbones. Because retrieval is semantic rather than genealogical, insights also transfer to problems and benchmark families from which they were not derived. The current store grows without bound, motivating consolidation and forgetting mechanisms at larger scales.


\section*{Limitations}

\sysname{} depends on LLM-generated insights and embedding-based retrieval, so inaccurate summaries, weak semantic matches, or noisy evaluation reports may store and reuse misleading guidance. Its effectiveness also depends on fixed clustering and retrieval hyperparameters, and the current memory grows without consolidation or forgetting, which may reduce efficiency and relevance at larger scales. Our comparisons match methods by the number of evaluated candidates rather than total LLM calls, tokens, wall-clock time, or monetary cost; because \sysname{} adds distillation, embedding, and recommendation-synthesis operations, reductions in evolution steps should not be interpreted as equivalent reductions in total compute. The main results use one run per model--harness pair because of the cost of full 100-step evolutionary trajectories, while the per-step, ablation, and transfer analyses use three seeds. Finally, we do not directly benchmark against alternative memory-based evolutionary systems, so our experiments establish the benefit of adding \sysname{} to matched memoryless harnesses rather than superiority over all existing memory mechanisms.

\section*{Ethical Considerations}
\label{sec:ethics}

\sysname{} inherits the biases, hallucinations, and unsupported reasoning of its underlying language models, and persistent memory may amplify an incorrect insight across later mutations or runs. The additional distillation, retrieval, and synthesis calls also increase computational and environmental cost, although retrieval is bounded by a fixed token budget. Our experiments use public benchmarks, models, and implementations under their respective licenses and involve no human subjects or newly collected personal data. We used LLM assistance only to improve manuscript presentation, with all technical decisions, experiments, and conclusions verified by the authors; released code and artifacts will document dependencies and will redistribute third-party data or model outputs only when permitted by their licenses.


\bibliography{custom}


\appendix

\section{Memory Construction and Retrieval Details}
\label{sec:memory_details}

\paragraph{Cluster assignment.}
Let $e_i\in\mathbb{R}^d$ denote the embedding of a newly distilled insight and let $c_k$ denote the representative of cluster $k$. We identify the nearest cluster as
\begin{equation}
k^*=\arg\max_k \cos(e_i,c_k).
\end{equation}
The insight joins $k^*$ when $\cos(e_i,c_{k^*})\geq\tau_{\mathrm{cluster}}$ and otherwise initializes a new cluster. Problem-level and change-level insights follow the same assignment procedure.

\paragraph{Attention-weighted representative.}
For a cluster containing embeddings $\{e_1,\ldots,e_n\}$, we compute the semantic centrality of member $i$ as
\begin{equation}
s_i=\frac{1}{n-1}\sum_{j\neq i}\cos(e_i,e_j).
\end{equation}
We transform these scores into non-parametric attention weights,
\begin{equation}
w_i=\frac{\exp(s_i/\tau)}{\sum_{j=1}^{n}\exp(s_j/\tau)},
\end{equation}
and form the cluster representative as
\begin{equation}
c=\sum_{i=1}^{n}w_i e_i.
\end{equation}
For a singleton cluster, we set $c=e_1$. As $\tau\rightarrow\infty$, the weights approach a uniform distribution and the representative approaches mean pooling.

\paragraph{Hierarchical retrieval.}
Given query embedding $e_q$, we rank clusters according to
\begin{equation}
r_k^{\mathrm{cluster}}=\cos(e_q,c_k)
\end{equation}
and retain the top $K_c$ clusters satisfying $r_k^{\mathrm{cluster}}\geq\tau_{\mathrm{ret}}$. Within each retained cluster, member insights are ranked by
\begin{equation}
r_i^{\mathrm{insight}}=\cos(e_q,e_i),
\end{equation}
and the top $K_i$ insights are selected. Cluster and insight similarities are retained separately rather than combined into a single score.

\paragraph{Evidence and recommendation synthesis.}
For each retrieved insight, \sysname{} optionally attaches up to two linked Level~1 traces. The synthesis prompt receives the current query, insight text, cluster similarity, insight similarity, and available raw evidence, and returns 3 actionable recommendations grounded in the retrieved records. If no relevant cluster is found, no synthesis call is made and the original mutation prompt is used unchanged.

\paragraph{Online updates.}
Memory construction and retrieval occur throughout a run. Each newly evaluated candidate immediately produces a Level~1 trace and a Level~2 insight, after which its embedding is assigned to Level~3. Because the store is persisted on disk, later mutations can retrieve insights written earlier in the same run as well as insights retained from previous runs.

\section{Experimental Setup Details}
\label{sec:setup_details}

\paragraph{Benchmarks.}
AlgoTune consists of numerical programming tasks in which a generated solver must reproduce a reference implementation while reducing execution time. We report the harmonic mean of per-task speedups. ALE-Bench contains AtCoder Heuristic Contest problems with continuous, non-saturating objectives. We use the ten-problem LITE subset and report average performance, a problem-agnostic score derived from the human leaderboard. Following \citet{imajuku2025alebench}, we avoid rating because it can be dominated by a single exceptional result, and additionally report the five short- and five long-format problems separately.

\paragraph{Evaluation protocol.}
During evolution, candidate fitness is computed only from the public local cases supplied by each benchmark. Final results are obtained through private evaluation, and private scores are never exposed to the search procedure. Because our 100-step ALE-Bench budget is smaller than that used in published leaderboard submissions, absolute scores should be interpreted only through matched comparisons within our experiments.

\paragraph{Harnesses and search budget.}
We attach \sysname{} to OpenEvolve and ShinkaEvolve through the evaluation-write and mutation-read hooks. Their original selection, island-based population management, evaluation, and mutation procedures remain unchanged. Each run contains at most 100 evolution steps, with one child generated and evaluated per step. The main tables report one run per model--harness pair because of the cost of full evolutionary trajectories. The per-step, component-ablation, embedding-ablation, and memory-transfer experiments are each averaged over three independent seeds.

\paragraph{Memory isolation.}
Memory is constructed from a disjoint set of problems before deployment on the evaluation set. No trace, insight, or cluster originating from an evaluation problem is available when that problem is first solved. During evaluation, the store is updated online, allowing later steps to retrieve insights generated earlier in the same trajectory together with eligible knowledge from prior runs.

\paragraph{Models and decoding.}
We evaluate GPT-5.4, DeepSeek V4 Pro, Gemini 3 Pro, Qwen 3.7 Max, and Qwen 3 Coder. Within each run, the same model performs mutation, trace distillation, optional cluster summarization, and recommendation synthesis, preventing memory-assisted conditions from receiving a stronger model than their corresponding baselines. We use a temperature of $0.7$ for mutation to preserve search diversity and $0.2$ for distillation, cluster summarization, and recommendation synthesis to improve consistency. We use \textsc{Qwen-Embedding-4B}~\citep{zhang2025qwen3embeddingadvancingtext} for all insight and query embeddings and keep it fixed across experiments.

\paragraph{Memory hyperparameters.}
We use a cluster-assignment threshold of $\tau_{\mathrm{cluster}}=0.80$, a retrieval threshold of $\tau_{\mathrm{ret}}=0.60$, and a centroid temperature of $\tau=0.10$. Retrieval expands at most $K_c=3$ clusters and selects at most $K_i=3$ insights from each cluster, under a fixed budget of 2,500 tokens. Cluster summaries use at most ten member insights, each retrieved insight may include up to two linked Level~1 traces, and the synthesizer returns exactly three recommendations. These settings are shared across benchmarks, models, and harnesses. If no cluster exceeds $\tau_{\mathrm{ret}}$, no recommendation-synthesis call is made and the original mutation prompt is used. Bounding retrieval by $K_c$, $K_i$, and the token budget keeps prompt length independent of the total memory size.

\section{ALE-Bench Results by Contest Format}
\label{sec:alebench_format}

Table~\ref{tab:alebench_format} separates ALE-Bench LITE performance into its five short- and five long-format problems. \sysname{} improves short-format performance by 8.1\% under OpenEvolve and 6.9\% under ShinkaEvolve, while improving long-format performance by 6.2\% and 5.1\%, respectively. The larger gains on short contests suggest that their recurring optimization strategies transfer more readily, while the consistent long-format improvements show that retrieved insights remain useful for more problem-specific search.

\begin{table}[t]
    \centering
    \footnotesize
    \caption{ALE-Bench LITE by contest format, averaged over models.}
    \label{tab:alebench_format}
    \begin{tabular}{lccc}
        \toprule
        Method & Short & Long & Overall \\
        \midrule
        OpenEvolve          & 1166.8 & 1021.8 & 1094.3 \\
        \quad + \sysname{}  & 1261.7 & 1085.5 & 1173.6 \\
        ShinkaEvolve        & 1209.4 & 1056.8 & 1133.1 \\
        \quad + \sysname{}  & 1292.6 & 1111.2 & 1201.9 \\
        \bottomrule
    \end{tabular}
\end{table}

\section{Qualitative Example of Memory Use} 
\label{sec:example}

This section illustrates how \sysname{} converts an evolutionary evaluation into reusable knowledge and later retrieves that knowledge to guide a concrete code modification. We consider a capacitated facility-location task in which an intermediate candidate achieves a $2.36\times$ speedup over the original reference implementation. A subsequent candidate replaces the CBC mixed-integer programming backend with OR-Tools CP-SAT and reaches a $3.99\times$ speedup. Because both values are measured relative to the same reference implementation, the CP-SAT candidate is $3.99/2.36=1.69\times$ faster than the intermediate candidate, corresponding to a further $40.9\%$ runtime reduction.

\subsection{Intermediate Candidate and Remaining Bottleneck}

The intermediate candidate formulates the problem using binary facility-opening variables $y_i$ and binary customer-assignment variables $x_{ij}$. Each customer is assigned to exactly one facility, and facility capacities are enforced through
\begin{equation}
    \sum_j d_j x_{ij} \leq C_i y_i.
\end{equation}
The core of the $2.36\times$ candidate is shown below.

\begin{sedimacode}{Intermediate candidate: CBC formulation at $2.36\times$}
from ortools.linear_solver import pywraplp

solver = pywraplp.Solver.CreateSolver("CBC")
if not solver:
    solver = pywraplp.Solver.CreateSolver("SCIP")

# Binary facility-opening and assignment variables
y = [
    solver.BoolVar(f"y_{i}")
    for i in range(n_facilities)
]

x = [
    [
        solver.BoolVar(f"x_{i}_{j}")
        for j in range(n_customers)
    ]
    for i in range(n_facilities)
]

# Objective
total_cost = sum(
    fixed_costs[i] * y[i]
    for i in range(n_facilities)
)
total_cost += sum(
    transportation_costs[i][j] * x[i][j]
    for i in range(n_facilities)
    for j in range(n_customers)
)
solver.Minimize(total_cost)

# Each customer is assigned to exactly one facility
for j in range(n_customers):
    solver.Add(
        sum(x[i][j] for i in range(n_facilities)) == 1
    )

# Facility-capacity constraints
for i in range(n_facilities):
    solver.Add(
        sum(
            demands[j] * x[i][j]
            for j in range(n_customers)
        ) <= capacities[i] * y[i]
    )

status = solver.Solve()
\end{sedimacode}

\begin{sedimabox}{Representative evaluator diagnosis}
The candidate is correct and improves substantially over the reference implementation, but its remaining runtime is dominated by solving a dense binary mixed-integer program with CBC. The model consists almost entirely of Boolean decisions and repeated exactly-one assignment constraints, yet these structures are represented using generic linear constraints. A Boolean-oriented solver with native cardinality constraints may exploit this structure more effectively.
\end{sedimabox}

The formulation contains $n_f n_c$ binary assignment variables in addition to the $n_f$ facility-opening variables. Python-side model construction and dense result extraction introduce some overhead, but the main remaining bottleneck is the CBC branch-and-bound search.

\subsection{Writing the Experience to \sysname{}}

After evaluation, \sysname{} stores the factual execution trace at Level~1. An LLM then converts the trace into a compact Level~2 insight describing the reusable optimization principle. The insight is embedded and assigned to a Level~3 cluster containing semantically related experiences from assignment, scheduling, routing, and resource-allocation tasks.

Table~\ref{tab:memory_snapshot} shows an example snapshot of the resulting memory state.
\subsection{Reading the Memory and Applying the Retrieved Insight}

Before the next mutation, \sysname{} embeds a query describing the current implementation and its remaining bottleneck. The query is close to the Boolean-assignment cluster because it contains the same structural signals as the stored experiences: binary assignment decisions, exactly-one constraints, capacity-limited resources, and a solver-dominated execution profile.

\begin{sedimabox}{Retrieved cross-problem experience}
A previous capacitated scheduling task used CBC even though nearly all decisions were Boolean. Recasting the model in CP-SAT and replacing generic sum-to-one equalities with native cardinality constraints substantially reduced solver time. Floating-point coefficients had to be converted to bounded integers before constructing the CP-SAT model.
\end{sedimabox}

The retrieved experience does not need to originate from the same facility-location task. Retrieval is based on semantic similarity, so an insight obtained from scheduling or resource assignment can transfer when the underlying computational pattern is similar.

The retrieved guidance results in the following candidate.

\begin{sedimacode}{Memory-guided candidate: CP-SAT formulation at $3.99\times$}
from ortools.sat.python import cp_model
import numpy as np

fixed_costs = np.asarray(problem["fixed_costs"])
capacities = np.asarray(problem["capacities"])
demands = np.asarray(problem["demands"])
transportation_costs = np.asarray(
    problem["transportation_costs"]
)

# CP-SAT requires integer coefficients
precision = 1_000_000
fixed_costs = (fixed_costs * precision).astype(int)
capacities = (capacities * precision).astype(int)
demands = (demands * precision).astype(int)
transportation_costs = (
    transportation_costs * precision
).astype(int)

model = cp_model.CpModel()

# Binary facility-opening and assignment variables
y = [
    model.NewBoolVar(f"y_{i}")
    for i in range(n_facilities)
]

x = {
    (i, j): model.NewBoolVar(f"x_{i}_{j}")
    for i in range(n_facilities)
    for j in range(n_customers)
}

# Native exactly-one constraints
for j in range(n_customers):
    model.AddExactlyOne(
        [x[i, j] for i in range(n_facilities)]
    )

# Facility-capacity constraints
for i in range(n_facilities):
    model.Add(
        sum(
            demands[j] * x[i, j]
            for j in range(n_customers)
        ) <= capacities[i] * y[i]
    )

# Integer-valued objective
objective = sum(
    fixed_costs[i] * y[i]
    for i in range(n_facilities)
)
objective += sum(
    transportation_costs[i][j] * x[i, j]
    for i in range(n_facilities)
    for j in range(n_customers)
)
model.Minimize(objective)

solver = cp_model.CpSolver()
solver.parameters.max_time_in_seconds = 30.0
status = solver.Solve(model)
\end{sedimacode}

\begin{table*}[t]
    \centering
    \footnotesize
    \setlength{\tabcolsep}{5pt}
    \caption{Example snapshot of the information written to and retrieved from \sysname{}. Level~1 preserves task-specific evidence, while Levels~2 and~3 support transfer across problems.}
    \label{tab:memory_snapshot}
    \begin{tabularx}{\textwidth}{@{}p{0.16\textwidth} X X@{}}
        \toprule
        Operation & Memory content & Function \\
        \midrule
        Level~1 write &
        \textbf{Task:} capacitated facility location. \textbf{Score:} $2.36\times$. \textbf{Backend:} CBC. \textbf{Structure:} dense binary assignment variables, facility-opening variables, exactly-one customer assignments, and capacity constraints. \textbf{Diagnosis:} solver execution remains the dominant cost. &
        Preserves the candidate score, implementation details, evaluator evidence, and provenance of the higher-level insight. \\
        \addlinespace
        Level~2 write &
        Boolean assignment models with exactly-one and capacity constraints may be solved more efficiently using a Boolean-oriented constraint solver. Native cardinality constraints should be exposed directly instead of being represented only as generic linear equalities. &
        Abstracts the facility-location trace into a reusable principle applicable to other discrete optimization problems. \\
        \addlinespace
        Level~3 write &
        \textbf{Cluster:} Boolean assignment models and solver selection. Related entries include job assignment, capacitated scheduling, routing, and resource-activation problems. &
        Groups experiences by computational structure rather than benchmark identity or evolutionary lineage. \\
        \addlinespace
        Retrieval query &
        The current program uses CBC for a dense Boolean facility-location model. Runtime is dominated by the solver, and each customer participates in an exactly-one assignment constraint. &
        Matches the stored cluster through shared concepts: Boolean decisions, assignment structure, capacities, and solver-bound execution. \\
        \addlinespace
        Retrieved guidance &
        Replace the generic MILP backend with CP-SAT, use native \texttt{AddExactlyOne} constraints, and convert floating-point coefficients to bounded integers. &
        Provides a concrete mutation direction that can be directly translated into code. \\
        \addlinespace
        Resulting candidate &
        CBC is replaced with \texttt{CpModel} and \texttt{CpSolver}; generic assignment equalities are replaced with \texttt{AddExactlyOne}; numerical coefficients are scaled to integers; and a bounded solve time is introduced. &
        Produces the $3.99\times$ candidate, a further $1.69\times$ improvement over the $2.36\times$ intermediate candidate. \\
        \bottomrule
    \end{tabularx}
\end{table*}

The retrieved memory leads to three principal modifications. First, the generic CBC mixed-integer solver is replaced with CP-SAT, which is designed for discrete models dominated by Boolean and integer variables. Second, the constraint $\sum_i x_{ij}=1$ is represented using the native \texttt{AddExactlyOne} primitive, exposing its cardinality structure directly to the solver. Third, floating-point costs, capacities, and demands are scaled to integers, as required by CP-SAT.

\begin{table*}[t]
    \centering
    \footnotesize
    \caption{Connection between the retrieved memory and the resulting code modifications.}
    \label{tab:memory_to_edit}
    \begin{tabularx}{\linewidth}{@{}X X@{}}
        \toprule
        Retrieved information & Resulting code modification \\
        \midrule
        Boolean-dominated optimization models may benefit from CP-SAT. &
        Replace \texttt{pywraplp.Solver} and CBC with \texttt{CpModel} and \texttt{CpSolver}. \\
        \addlinespace
        Exactly-one assignment structure should be exposed natively. &
        Replace generic sum-equals-one constraints with \texttt{AddExactlyOne}. \\
        \addlinespace
        CP-SAT requires bounded integer coefficients. &
        Scale costs, demands, and capacities by a shared precision factor and convert them to integers. \\
        \addlinespace
        Difficult solver runs should have bounded execution time. &
        Set \texttt{max\_time\_in\_seconds} to $30$. \\
        \bottomrule
    \end{tabularx}
\end{table*}

\subsection{Interpretation}

This example demonstrates the full \sysname{} write--retrieve--apply cycle. The evaluator first identifies that the $2.36\times$ candidate remains solver-bound. \sysname{} stores the raw evidence, distills it into a solver-selection insight, and clusters it with structurally similar experiences. During a later mutation, the current program retrieves this insight and uses it to reformulate the model with CP-SAT, producing the $3.99\times$ candidate.

The $3.99\times$ implementation also introduces a $30$-second solver limit and accepts both \texttt{OPTIMAL} and \texttt{FEASIBLE} CP-SAT solutions, whereas the CBC implementation returns a solution only after proving optimality. The observed gain may therefore reflect both a more suitable solver formulation and a different termination policy. A controlled evaluation should additionally compare objective values and report how often CP-SAT terminates with an optimal rather than merely feasible solution.

\section{Ablation Studies}
\label{sec:ablations}

We conduct all ablations under OpenEvolve using three independent seeds for each of the five mutation models and report results averaged across both seeds and models. To control evaluation cost, AlgoTune results use its first 20 tasks, while ALE-Bench results use the complete ten-problem LITE subset. The full-\sysname{} and no-memory ALE-Bench means happen to round to the same values as the corresponding OpenEvolve means in Table~\ref{tab:alebench}, but are computed from these separate three-seed ablation runs.

\paragraph{Memory components.}
Table~\ref{tab:ablation} isolates the contribution of each memory component. The full \sysname{} achieves the strongest performance on both benchmarks, improving over no memory by 5.0\% on AlgoTune-20 and 7.2\% on ALE-Bench LITE. Replacing the attention-weighted representative with a mean centroid causes a modest decline, indicating that emphasizing prototypical insights improves retrieval. Using Level~2 insights without semantic clustering produces a larger drop, showing the benefit of hierarchical organization. Random retrieval performs only marginally above no memory, confirming that gains arise from retrieving relevant experience rather than merely adding prior text to the mutation prompt.

\begin{table}[t]
    \centering
    \small
    \caption{Component ablation under OpenEvolve, averaged over three seeds and five mutation models. AlgoTune results use the first 20 tasks, while ALE-Bench uses the full LITE set.}
    \label{tab:ablation}
    
    \begin{tabular}{lcc}
        \toprule
        Variant & AlgoTune-20 & ALE-Bench LITE \\
        \midrule
        \sysname{} (full)      & 1.69 & 1173.6 \\
        \quad Level 2 only     & 1.65 & 1138.9 \\
        \quad Mean centroid    & 1.67 & 1157.4 \\
        \quad Random retrieval & 1.62 & 1102.6 \\
        \quad No memory        & 1.61 & 1094.3 \\
        \bottomrule
    \end{tabular}
\end{table}

\paragraph{Embedding model.}
Table~\ref{tab:embedding_ablation} evaluates whether \sysname{} depends on a particular embedding backbone. Qwen3-Embedding-4B performs best on both benchmarks, but Qwen3-Embedding-8B and GTE-Qwen2-7B-Instruct remain close, indicating that the method is robust across strong semantic encoders. Performance declines more noticeably with BGE-M3 and Qwen3-Embedding-0.6B, suggesting that weaker representations reduce cluster assignment and retrieval quality. Nevertheless, every embedding-based configuration outperforms the no-memory baseline, showing that the benefit of \sysname{} is not specific to its default encoder.

\begin{table}[t]
    \centering
    \small
    \caption{Embedding-model ablation under OpenEvolve, averaged over three seeds and five mutation models. AlgoTune results use the first 20 tasks, while ALE-Bench uses the full LITE set.}
    \label{tab:embedding_ablation}
    \resizebox{\linewidth}{!}{%
    \begin{tabular}{lcc}
        \toprule
        Embedding model & AlgoTune-20 & ALE-Bench LITE \\
        \midrule
        Qwen3-Embedding-4B~\citeyear{zhang2025qwen3embeddingadvancingtext} & 1.69 & 1173.6 \\
        Qwen3-Embedding-8B~\citeyear{zhang2025qwen3embeddingadvancingtext}           & 1.68 & 1168.4 \\
        GTE-Qwen2-7B-Instruct~\citeyear{li2023generaltextembeddingsmultistage}        & 1.67 & 1159.7 \\
        BGE-M3~\citeyear{chen-etal-2024-m3}                       & 1.65 & 1147.2 \\
        Qwen3-Embedding-0.6B~\citeyear{zhang2025qwen3embeddingadvancingtext}         & 1.64 & 1136.8 \\
        \bottomrule
    \end{tabular}
    }
\end{table}

\section{Per-Problem Results}
\label{sec:per_problem_performance}

\subsection{AlgoTune}
\label{sec:algotune_per_problem_performance}

Tables~\ref{tab:algotune_per_task_1}, \ref{tab:algotune_per_task_2}, \ref{tab:algotune_per_task_shinka_1}, and \ref{tab:algotune_per_task_shinka_2} report the full AlgoTune breakdown. Performance is strongly task dependent: many low-level kernels remain near $1\times$, whereas graph, optimization, control, and numerical tasks such as \texttt{lp\_mdp}, \texttt{water\_filling}, and \texttt{convolve2d\_full\_fill} admit substantially larger speedups. ShinkaEvolve generally achieves slightly stronger per-task results than OpenEvolve, while the best mutation model varies across tasks, showing that the aggregate gains are not driven by a single model or problem category.

\begin{table*}[t]
\centering
\scriptsize
\setlength{\tabcolsep}{5pt}
\caption{Per-task speedup on AlgoTune using OpenEvolve+\sysname{} (part 1 of 2).}
\label{tab:algotune_per_task_1}
\begin{tabular}{@{}lrrrrr@{}}
\toprule
Task & GPT-5.4 & DeepSeek V4 Pro & Gemini 3 Pro & Qwen 3.7 Max & Qwen 3 Coder \\
\midrule
\texttt{aes\_gcm\_encryption} & 1.12 & 1.06 & 1.18 & 1.05 & 1.02 \\
\texttt{affine\_transform\_2d} & 1.00 & 1.00 & 1.00 & 1.00 & 1.00 \\
\texttt{aircraft\_wing\_design} & 1.08 & 1.12 & 2.13 & 1.21 & 1.14 \\
\texttt{articulation\_points} & 2.37 & 4.95 & 2.39 & 3.87 & 2.77 \\
\texttt{base64\_encoding} & 1.06 & 1.00 & 1.10 & 1.00 & 1.00 \\
\texttt{battery\_scheduling} & 8.54 & 14.85 & 31.72 & 15.62 & 12.94 \\
\texttt{btsp} & 1.19 & 2.39 & 2.34 & 2.00 & 1.30 \\
\texttt{capacitated\_facility\_location} & 4.58 & 10.34 & 14.01 & 7.37 & 2.45 \\
\texttt{chacha\_encryption} & 1.22 & 1.18 & 1.24 & 1.16 & 1.08 \\
\texttt{channel\_capacity} & 1.07 & 1.14 & 1.45 & 1.15 & 1.10 \\
\texttt{chebyshev\_center} & 2.99 & 4.58 & 7.98 & 4.63 & 3.91 \\
\texttt{cholesky\_factorization} & 1.07 & 1.04 & 1.08 & 1.04 & 1.02 \\
\texttt{clustering\_outliers} & 1.16 & 1.29 & 1.20 & 1.25 & 1.17 \\
\texttt{communicability} & 23.50 & 54.32 & 146.67 & 59.25 & 50.18 \\
\texttt{convex\_hull} & 1.41 & 3.83 & 6.02 & 3.19 & 1.65 \\
\texttt{convolve2d\_full\_fill} & 46.10 & 99.73 & 174.44 & 95.49 & 68.89 \\
\texttt{convolve\_1d} & 1.04 & 1.10 & 1.13 & 1.08 & 1.03 \\
\texttt{correlate2d\_full\_fill} & 34.41 & 100.21 & 133.94 & 90.74 & 63.02 \\
\texttt{correlate\_1d} & 1.05 & 1.08 & 1.25 & 1.08 & 1.04 \\
\texttt{count\_connected\_components} & 2.49 & 5.44 & 5.07 & 4.71 & 3.48 \\
\texttt{count\_riemann\_zeta\_zeros} & 1.00 & 1.00 & 1.00 & 1.00 & 1.00 \\
\texttt{cumulative\_simpson\_1d} & 5.49 & 9.37 & 12.95 & 6.89 & 2.42 \\
\texttt{cumulative\_simpson\_multid} & 1.03 & 1.00 & 1.05 & 1.00 & 1.00 \\
\texttt{cvar\_projection} & 1.37 & 2.65 & 1.52 & 2.23 & 1.73 \\
\texttt{cyclic\_independent\_set} & 1.47 & 10.57 & 2.12 & 5.43 & 1.76 \\
\texttt{dct\_type\_I\_scipy\_fftpack} & 1.06 & 1.08 & 1.45 & 1.11 & 1.06 \\
\texttt{delaunay} & 1.85 & 3.29 & 1.95 & 2.50 & 1.43 \\
\texttt{dijkstra\_from\_indices} & 1.00 & 1.00 & 1.00 & 1.00 & 1.00 \\
\texttt{discrete\_log} & 1.04 & 1.35 & 1.11 & 1.24 & 1.07 \\
\texttt{dst\_type\_II\_scipy\_fftpack} & 1.28 & 1.63 & 1.33 & 1.47 & 1.17 \\
\texttt{dynamic\_assortment\_planning} & 19.75 & 32.40 & 47.10 & 22.44 & 5.23 \\
\texttt{earth\_movers\_distance} & 1.01 & 1.01 & 1.01 & 1.01 & 1.00 \\
\texttt{edge\_expansion} & 3.90 & 13.45 & 3.12 & 7.35 & 2.27 \\
\texttt{eigenvalues\_complex} & 1.32 & 1.70 & 2.15 & 1.66 & 1.49 \\
\texttt{eigenvalues\_real} & 1.85 & 2.90 & 4.04 & 2.79 & 2.32 \\
\texttt{eigenvectors\_complex} & 1.03 & 1.05 & 1.06 & 1.04 & 1.02 \\
\texttt{eigenvectors\_real} & 1.02 & 1.04 & 1.12 & 1.04 & 1.03 \\
\texttt{elementwise\_integration} & 1.00 & 1.00 & 1.00 & 1.00 & 1.00 \\
\texttt{feedback\_controller\_design} & 21.53 & 102.99 & 79.46 & 53.35 & 6.32 \\
\texttt{fft\_cmplx\_scipy\_fftpack} & 1.71 & 2.65 & 2.57 & 2.48 & 2.22 \\
\texttt{fft\_convolution} & 1.00 & 1.00 & 1.00 & 1.00 & 1.00 \\
\texttt{fft\_real\_scipy\_fftpack} & 1.08 & 1.67 & 1.29 & 1.79 & 2.35 \\
\texttt{firls} & 1.00 & 1.00 & 1.00 & 1.01 & 1.01 \\
\texttt{generalized\_eigenvalues\_complex} & 2.50 & 4.76 & 4.23 & 4.07 & 3.05 \\
\texttt{generalized\_eigenvalues\_real} & 1.86 & 3.27 & 3.92 & 3.00 & 2.35 \\
\texttt{generalized\_eigenvectors\_complex} & 1.68 & 3.21 & 4.14 & 2.74 & 1.57 \\
\texttt{generalized\_eigenvectors\_real} & 1.41 & 2.09 & 4.20 & 2.20 & 1.95 \\
\texttt{graph\_coloring\_assign} & 3.44 & 2.68 & 2.54 & 2.44 & 1.60 \\
\texttt{graph\_global\_efficiency} & 2.87 & 10.11 & 18.73 & 9.67 & 8.08 \\
\texttt{graph\_isomorphism} & 19.01 & 44.19 & 65.78 & 37.46 & 21.06 \\
\texttt{graph\_laplacian} & 1.00 & 1.00 & 1.00 & 1.00 & 1.00 \\
\texttt{group\_lasso} & 1.00 & 1.00 & 1.00 & 1.00 & 1.00 \\
\texttt{gzip\_compression} & 1.12 & 1.10 & 1.11 & 1.09 & 1.05 \\
\texttt{integer\_factorization} & 1.00 & 1.00 & 1.00 & 1.00 & 1.00 \\
\texttt{job\_shop\_scheduling} & 1.43 & 1.95 & 2.17 & 1.75 & 1.30 \\
\texttt{kalman\_filter} & 5.72 & 12.54 & 14.94 & 9.17 & 2.85 \\
\texttt{kcenters} & 1.44 & 2.03 & 1.49 & 1.74 & 1.25 \\
\texttt{kd\_tree} & 1.06 & 1.07 & 1.06 & 1.06 & 1.03 \\
\texttt{kernel\_density\_estimation} & 1.00 & 1.00 & 1.00 & 1.00 & 1.00 \\
\texttt{kmeans} & 6.30 & 12.69 & 21.17 & 12.59 & 10.18 \\
\texttt{ks\_test\_2samp} & 1.03 & 1.00 & 1.05 & 1.00 & 1.00 \\
\texttt{l0\_pruning} & 1.23 & 1.74 & 3.26 & 1.94 & 2.12 \\
\texttt{l1\_pruning} & 2.84 & 2.92 & 3.57 & 2.76 & 1.92 \\
\texttt{lasso} & 1.12 & 1.59 & 1.21 & 1.41 & 1.12 \\
\texttt{least\_squares} & 1.16 & 2.25 & 1.89 & 2.05 & 1.81 \\
\texttt{linear\_system\_solver} & 1.09 & 1.15 & 1.30 & 1.14 & 1.09 \\
\texttt{lp\_box} & 6.54 & 14.39 & 22.13 & 13.38 & 9.63 \\
\texttt{lp\_centering} & 1.01 & 1.02 & 1.01 & 1.02 & 1.01 \\
\texttt{lp\_mdp} & 87.53 & 163.16 & 291.43 & 110.55 & 21.50 \\
\texttt{lqr} & 1.08 & 1.32 & 1.58 & 1.27 & 1.10 \\
\texttt{lti\_simulation} & 1.51 & 6.99 & 3.63 & 4.43 & 1.73 \\
\texttt{lu\_factorization} & 1.00 & 1.05 & 1.01 & 1.08 & 1.15 \\
\texttt{lyapunov\_stability} & 38.44 & 101.27 & 109.17 & 86.73 & 58.43 \\
\texttt{markowitz} & 1.01 & 1.00 & 1.02 & 1.00 & 1.00 \\
\texttt{matrix\_completion} & 1.00 & 1.01 & 1.01 & 1.01 & 1.02 \\
\texttt{matrix\_exponential} & 1.00 & 1.00 & 1.00 & 1.00 & 1.00 \\
\texttt{matrix\_exponential\_sparse} & 1.00 & 1.00 & 1.00 & 1.00 & 1.00 \\
\bottomrule
\end{tabular}
\end{table*}

\begin{table*}[t]
\centering
\scriptsize
\setlength{\tabcolsep}{5pt}
\caption{Per-task speedup on AlgoTune using OpenEvolve+\sysname{} (part 2 of 2).}
\label{tab:algotune_per_task_2}
\begin{tabular}{@{}lrrrrr@{}}
\toprule
Task & GPT-5.4 & DeepSeek V4 Pro & Gemini 3 Pro & Qwen 3.7 Max & Qwen 3 Coder \\
\midrule
\texttt{matrix\_multiplication} & 1.03 & 1.02 & 1.03 & 1.02 & 1.01 \\
\texttt{matrix\_sqrt} & 1.02 & 1.02 & 1.02 & 1.02 & 1.01 \\
\texttt{max\_clique\_cpsat} & 4.67 & 12.24 & 13.52 & 10.57 & 6.01 \\
\texttt{max\_common\_subgraph} & 4.92 & 4.19 & 12.34 & 4.38 & 2.49 \\
\texttt{max\_flow\_min\_cost} & 2.68 & 6.90 & 4.03 & 5.99 & 6.21 \\
\texttt{max\_independent\_set\_cpsat} & 4.54 & 3.18 & 3.31 & 2.87 & 1.74 \\
\texttt{max\_weighted\_independent\_set} & 1.20 & 1.18 & 1.63 & 1.21 & 1.12 \\
\texttt{min\_dominating\_set} & 1.08 & 1.83 & 1.21 & 1.65 & 1.48 \\
\texttt{min\_weight\_assignment} & 1.16 & 1.64 & 1.32 & 1.43 & 1.12 \\
\texttt{minimum\_spanning\_tree} & 3.76 & 6.24 & 3.17 & 4.11 & 1.76 \\
\texttt{minimum\_volume\_ellipsoid} & 3.53 & 18.21 & 18.53 & 11.59 & 2.93 \\
\texttt{multi\_dim\_knapsack} & 2.60 & 16.97 & 9.41 & 9.83 & 2.94 \\
\texttt{nmf} & 1.03 & 1.23 & 1.13 & 1.21 & 1.20 \\
\texttt{ode\_brusselator} & 6.42 & 3.67 & 2.95 & 3.26 & 1.89 \\
\texttt{ode\_fitzhughnagumo} & 1.05 & 1.12 & 1.09 & 1.08 & 1.03 \\
\texttt{ode\_hires} & 5.69 & 17.23 & 29.18 & 13.76 & 5.76 \\
\texttt{ode\_hodgkinhuxley} & 19.23 & 40.29 & 91.74 & 33.97 & 11.25 \\
\texttt{ode\_lorenz96\_nonchaotic} & 1.52 & 2.79 & 2.92 & 2.47 & 1.84 \\
\texttt{ode\_lotkavolterra} & 16.93 & 11.91 & 48.16 & 12.96 & 5.58 \\
\texttt{ode\_nbodyproblem} & 16.06 & 32.05 & 33.01 & 25.60 & 14.15 \\
\texttt{ode\_seirs} & 24.48 & 6.58 & 43.39 & 10.82 & 11.96 \\
\texttt{ode\_stiff\_robertson} & 4.34 & 22.37 & 6.37 & 12.03 & 3.75 \\
\texttt{ode\_stiff\_vanderpol} & 24.79 & 45.97 & 6.12 & 21.65 & 4.88 \\
\texttt{odr} & 1.01 & 1.02 & 1.01 & 1.02 & 1.01 \\
\texttt{optimal\_advertising} & 1.04 & 1.31 & 1.10 & 1.22 & 1.06 \\
\texttt{outer\_product} & 1.13 & 1.03 & 1.21 & 1.02 & 1.01 \\
\texttt{pagerank} & 1.45 & 1.97 & 16.91 & 3.03 & 3.46 \\
\texttt{pca} & 2.14 & 3.75 & 4.06 & 3.01 & 1.61 \\
\texttt{pde\_burgers1d} & 1.34 & 3.74 & 4.56 & 3.55 & 3.07 \\
\texttt{pde\_heat1d} & 1.30 & 1.96 & 1.40 & 1.67 & 1.21 \\
\texttt{polynomial\_mixed} & 4.96 & 5.55 & 2.93 & 4.16 & 1.96 \\
\texttt{polynomial\_real} & 6.57 & 35.49 & 4.16 & 14.86 & 2.90 \\
\texttt{power\_control} & 22.66 & 139.92 & 122.11 & 97.36 & 27.97 \\
\texttt{procrustes} & 1.32 & 1.28 & 1.29 & 1.26 & 1.13 \\
\texttt{psd\_cone\_projection} & 3.15 & 4.65 & 10.56 & 5.38 & 5.56 \\
\texttt{qp} & 1.38 & 1.90 & 2.56 & 1.87 & 1.68 \\
\texttt{qr\_factorization} & 1.90 & 1.62 & 1.80 & 1.63 & 1.41 \\
\texttt{quantile\_regression} & 1.16 & 1.34 & 1.88 & 1.35 & 1.24 \\
\texttt{queens\_with\_obstacles} & 1.67 & 3.18 & 4.26 & 2.92 & 2.05 \\
\texttt{queuing} & 1.06 & 1.04 & 1.07 & 1.04 & 1.02 \\
\texttt{qz\_factorization} & 1.21 & 1.17 & 1.19 & 1.16 & 1.08 \\
\texttt{randomized\_svd} & 2.04 & 3.70 & 2.14 & 2.73 & 1.47 \\
\texttt{rbf\_interpolation} & 1.01 & 1.01 & 1.01 & 1.01 & 1.00 \\
\texttt{rectanglepacking} & 1.33 & 1.33 & 3.03 & 1.46 & 1.25 \\
\texttt{robust\_kalman\_filter} & 2.43 & 3.45 & 9.61 & 3.97 & 3.43 \\
\texttt{robust\_linear\_program} & 2.14 & 1.88 & 7.41 & 2.23 & 1.67 \\
\texttt{rocket\_landing\_optimization} & 1.07 & 1.59 & 1.24 & 1.40 & 1.11 \\
\texttt{rotate\_2d} & 1.00 & 1.00 & 1.00 & 1.00 & 1.00 \\
\texttt{set\_cover} & 4.05 & 6.38 & 4.48 & 4.65 & 2.00 \\
\texttt{set\_cover\_conflicts} & 1.95 & 2.72 & 3.44 & 2.69 & 2.24 \\
\texttt{sha256\_hashing} & 1.00 & 1.00 & 1.00 & 1.00 & 1.00 \\
\texttt{shift\_2d} & 1.00 & 1.00 & 1.00 & 1.00 & 1.00 \\
\texttt{shortest\_path\_dijkstra} & 1.71 & 2.50 & 3.87 & 2.25 & 1.45 \\
\texttt{sinkhorn} & 1.36 & 2.05 & 3.14 & 1.92 & 1.35 \\
\texttt{sparse\_eigenvectors\_complex} & 1.00 & 1.00 & 1.00 & 1.00 & 1.00 \\
\texttt{sparse\_lowest\_eigenvalues\_posdef} & 1.47 & 2.13 & 2.56 & 2.04 & 1.76 \\
\texttt{sparse\_lowest\_eigenvectors\_posdef} & 1.52 & 2.55 & 2.82 & 2.29 & 1.71 \\
\texttt{sparse\_pca} & 2.65 & 6.86 & 8.32 & 5.45 & 2.60 \\
\texttt{spectral\_clustering} & 3.99 & 8.57 & 14.73 & 6.43 & 2.31 \\
\texttt{stable\_matching} & 1.37 & 1.83 & 2.36 & 1.79 & 1.57 \\
\texttt{svd} & 1.19 & 1.17 & 1.18 & 1.16 & 1.08 \\
\texttt{svm} & 1.00 & 1.00 & 1.00 & 1.00 & 1.00 \\
\texttt{sylvester\_solver} & 1.02 & 1.01 & 1.02 & 1.01 & 1.01 \\
\texttt{tensor\_completion\_3d} & 14.09 & 23.09 & 40.25 & 18.72 & 6.10 \\
\texttt{toeplitz\_solver} & 1.00 & 1.00 & 1.00 & 1.00 & 1.00 \\
\texttt{tsp} & 1.16 & 1.37 & 1.29 & 1.29 & 1.19 \\
\texttt{two\_eigenvalues\_around\_0} & 1.52 & 2.04 & 2.77 & 1.99 & 1.72 \\
\texttt{unit\_simplex\_projection} & 1.71 & 3.18 & 1.80 & 2.44 & 1.42 \\
\texttt{upfirdn1d} & 1.02 & 1.16 & 1.05 & 1.11 & 1.03 \\
\texttt{vector\_quantization} & 1.00 & 1.00 & 1.00 & 1.01 & 1.01 \\
\texttt{vectorized\_newton} & 1.00 & 1.00 & 1.00 & 1.00 & 1.00 \\
\texttt{vehicle\_routing} & 1.22 & 1.54 & 3.44 & 1.69 & 1.53 \\
\texttt{vertex\_cover} & 1.39 & 1.62 & 3.43 & 1.68 & 1.31 \\
\texttt{voronoi\_diagram} & 2.24 & 2.13 & 4.91 & 2.49 & 2.35 \\
\texttt{wasserstein\_dist} & 4.48 & 8.48 & 9.47 & 7.16 & 4.52 \\
\texttt{water\_filling} & 60.54 & 86.86 & 195.98 & 95.57 & 82.87 \\
\texttt{zoom\_2d} & 1.00 & 1.00 & 1.00 & 1.00 & 1.00 \\
\bottomrule
\end{tabular}
\end{table*}

\begin{table*}[t]
\centering
\scriptsize
\setlength{\tabcolsep}{3pt}
\caption{Per-task speedup on AlgoTune using ShinkaEvolve+\sysname{} (part 1 of 2).}
\label{tab:algotune_per_task_shinka_1}
\begin{tabular}{@{}lrrrrr@{}}
\toprule
Task & GPT-5.4 & DeepSeek V4 Pro & Gemini 3 Pro & Qwen 3.7 Max & Qwen 3 Coder \\
\midrule
\texttt{aes\_gcm\_encryption} & 1.13 & 1.07 & 1.19 & 1.06 & 1.02 \\

\texttt{affine\_transform\_2d} & 1.00 & 1.00 & 1.00 & 1.00 & 1.00 \\
\texttt{aircraft\_wing\_design} & 1.09 & 1.13 & 2.16 & 1.22 & 1.16 \\
\texttt{articulation\_points} & 2.66 & 5.17 & 2.53 & 4.04 & 2.91 \\
\texttt{base64\_encoding} & 1.07 & 1.00 & 1.11 & 1.00 & 1.00 \\
\texttt{battery\_scheduling} & 9.85 & 16.16 & 34.92 & 17.21 & 14.44 \\
\texttt{btsp} & 1.21 & 2.62 & 2.41 & 2.11 & 1.36 \\
\texttt{capacitated\_facility\_location} & 5.34 & 10.95 & 14.87 & 8.11 & 2.57 \\
\texttt{chacha\_encryption} & 1.26 & 1.21 & 1.25 & 1.17 & 1.09 \\
\texttt{channel\_capacity} & 1.08 & 1.16 & 1.48 & 1.17 & 1.12 \\
\texttt{chebyshev\_center} & 3.14 & 4.75 & 8.46 & 5.26 & 4.26 \\
\texttt{cholesky\_factorization} & 1.08 & 1.04 & 1.08 & 1.04 & 1.02 \\
\texttt{clustering\_outliers} & 1.19 & 1.32 & 1.20 & 1.28 & 1.20 \\
\texttt{communicability} & 26.14 & 59.51 & 159.87 & 67.34 & 57.50 \\
\texttt{convex\_hull} & 1.45 & 3.94 & 6.03 & 3.56 & 1.79 \\
\texttt{convolve2d\_full\_fill} & 54.19 & 113.56 & 185.31 & 108.58 & 81.33 \\
\texttt{convolve\_1d} & 1.05 & 1.12 & 1.14 & 1.09 & 1.04 \\
\texttt{correlate2d\_full\_fill} & 37.90 & 111.36 & 140.01 & 104.27 & 72.26 \\
\texttt{correlate\_1d} & 1.06 & 1.09 & 1.27 & 1.09 & 1.05 \\
\texttt{count\_connected\_components} & 2.77 & 6.18 & 5.34 & 5.02 & 3.88 \\
\texttt{count\_riemann\_zeta\_zeros} & 1.00 & 1.00 & 1.00 & 1.00 & 1.00 \\
\texttt{cumulative\_simpson\_1d} & 6.34 & 10.97 & 13.73 & 7.49 & 2.71 \\
\texttt{cumulative\_simpson\_multid} & 1.03 & 1.00 & 1.05 & 1.00 & 1.00 \\
\texttt{cvar\_projection} & 1.44 & 2.79 & 1.58 & 2.28 & 1.82 \\
\texttt{cyclic\_independent\_set} & 1.51 & 11.99 & 2.28 & 5.69 & 1.92 \\
\texttt{dct\_type\_I\_scipy\_fftpack} & 1.07 & 1.08 & 1.50 & 1.12 & 1.07 \\
\texttt{delaunay} & 1.97 & 3.59 & 2.02 & 2.70 & 1.50 \\
\texttt{dijkstra\_from\_indices} & 1.00 & 1.00 & 1.00 & 1.00 & 1.00 \\
\texttt{discrete\_log} & 1.04 & 1.40 & 1.11 & 1.27 & 1.08 \\
\texttt{dst\_type\_II\_scipy\_fftpack} & 1.34 & 1.71 & 1.33 & 1.50 & 1.20 \\
\texttt{dynamic\_assortment\_planning} & 22.07 & 36.89 & 49.50 & 24.26 & 5.69 \\
\texttt{earth\_movers\_distance} & 1.01 & 1.01 & 1.01 & 1.01 & 1.00 \\
\texttt{edge\_expansion} & 4.36 & 15.20 & 3.32 & 7.84 & 2.46 \\
\texttt{eigenvalues\_complex} & 1.37 & 1.79 & 2.28 & 1.77 & 1.55 \\
\texttt{eigenvalues\_real} & 1.98 & 3.11 & 4.05 & 3.04 & 2.59 \\
\texttt{eigenvectors\_complex} & 1.04 & 1.06 & 1.06 & 1.05 & 1.02 \\
\texttt{eigenvectors\_real} & 1.02 & 1.04 & 1.13 & 1.05 & 1.04 \\
\texttt{elementwise\_integration} & 1.00 & 1.00 & 1.00 & 1.00 & 1.00 \\
\texttt{feedback\_controller\_design} & 23.81 & 118.57 & 83.39 & 61.71 & 6.82 \\
\texttt{fft\_cmplx\_scipy\_fftpack} & 1.76 & 2.94 & 2.65 & 2.61 & 2.33 \\
\texttt{fft\_convolution} & 1.00 & 1.00 & 1.00 & 1.00 & 1.00 \\
\texttt{fft\_real\_scipy\_fftpack} & 1.10 & 1.71 & 1.30 & 1.90 & 2.57 \\
\texttt{firls} & 1.00 & 1.00 & 1.00 & 1.01 & 1.01 \\
\texttt{generalized\_eigenvalues\_complex} & 2.85 & 5.06 & 4.69 & 4.39 & 3.23 \\
\texttt{generalized\_eigenvalues\_real} & 2.00 & 3.35 & 4.13 & 3.36 & 2.56 \\
\texttt{generalized\_eigenvectors\_complex} & 1.79 & 3.56 & 4.43 & 2.87 & 1.62 \\
\texttt{generalized\_eigenvectors\_real} & 1.45 & 2.28 & 4.35 & 2.24 & 2.14 \\
\texttt{graph\_coloring\_assign} & 3.71 & 2.78 & 2.58 & 2.59 & 1.71 \\
\texttt{graph\_global\_efficiency} & 3.05 & 11.17 & 20.36 & 11.33 & 9.56 \\
\texttt{graph\_isomorphism} & 21.29 & 50.99 & 67.89 & 42.08 & 23.94 \\
\texttt{graph\_laplacian} & 1.00 & 1.00 & 1.00 & 1.00 & 1.00 \\
\texttt{group\_lasso} & 1.00 & 1.00 & 1.00 & 1.00 & 1.00 \\
\texttt{gzip\_compression} & 1.13 & 1.11 & 1.12 & 1.10 & 1.06 \\
\texttt{integer\_factorization} & 1.00 & 1.00 & 1.00 & 1.00 & 1.00 \\
\texttt{job\_shop\_scheduling} & 1.46 & 2.03 & 2.22 & 1.81 & 1.32 \\
\texttt{kalman\_filter} & 6.66 & 13.52 & 16.43 & 10.01 & 3.12 \\
\texttt{kcenters} & 1.49 & 2.10 & 1.50 & 1.85 & 1.29 \\
\texttt{kd\_tree} & 1.07 & 1.08 & 1.06 & 1.06 & 1.04 \\
\texttt{kernel\_density\_estimation} & 1.00 & 1.00 & 1.00 & 1.00 & 1.00 \\
\texttt{kmeans} & 6.79 & 14.33 & 23.46 & 13.54 & 11.07 \\
\texttt{ks\_test\_2samp} & 1.03 & 1.00 & 1.06 & 1.00 & 1.00 \\
\texttt{l0\_pruning} & 1.27 & 1.84 & 3.45 & 2.01 & 2.37 \\
\texttt{l1\_pruning} & 3.20 & 3.01 & 3.87 & 2.82 & 2.08 \\
\texttt{lasso} & 1.14 & 1.64 & 1.24 & 1.48 & 1.14 \\
\texttt{least\_squares} & 1.17 & 2.45 & 1.91 & 2.16 & 1.98 \\
\texttt{linear\_system\_solver} & 1.11 & 1.16 & 1.34 & 1.15 & 1.11 \\
\texttt{lp\_box} & 7.15 & 15.92 & 23.28 & 14.17 & 11.54 \\
\texttt{lp\_centering} & 1.01 & 1.02 & 1.01 & 1.02 & 1.01 \\
\texttt{lp\_mdp} & 96.46 & 181.89 & 320.19 & 123.88 & 24.08 \\
\texttt{lqr} & 1.09 & 1.37 & 1.60 & 1.28 & 1.11 \\
\texttt{lti\_simulation} & 1.57 & 7.37 & 3.74 & 4.89 & 1.86 \\
\texttt{lu\_factorization} & 1.00 & 1.06 & 1.01 & 1.10 & 1.18 \\
\texttt{lyapunov\_stability} & 44.91 & 116.73 & 116.92 & 94.38 & 64.85 \\
\texttt{markowitz} & 1.01 & 1.00 & 1.02 & 1.00 & 1.00 \\
\texttt{matrix\_completion} & 1.00 & 1.01 & 1.01 & 1.01 & 1.02 \\
\texttt{matrix\_exponential} & 1.00 & 1.00 & 1.00 & 1.00 & 1.00 \\
\texttt{matrix\_exponential\_sparse} & 1.00 & 1.00 & 1.00 & 1.00 & 1.00 \\
\bottomrule
\end{tabular}
\end{table*}

\begin{table*}[t]
\centering
\scriptsize
\setlength{\tabcolsep}{3pt}
\caption{Per-task speedup on AlgoTune using ShinkaEvolve+\sysname{} (part 2 of 2).}
\label{tab:algotune_per_task_shinka_2}
\begin{tabular}{@{}lrrrrr@{}}
\toprule
Task & GPT-5.4 & DeepSeek V4 Pro & Gemini 3 Pro & Qwen 3.7 Max & Qwen 3 Coder \\
\midrule
\texttt{matrix\_multiplication} & 1.04 & 1.02 & 1.03 & 1.02 & 1.01 \\
\texttt{matrix\_sqrt} & 1.02 & 1.02 & 1.02 & 1.02 & 1.01 \\
\texttt{max\_clique\_cpsat} & 5.37 & 13.14 & 13.90 & 11.77 & 6.93 \\
\texttt{max\_common\_subgraph} & 5.64 & 4.52 & 13.20 & 5.00 & 2.67 \\
\texttt{max\_flow\_min\_cost} & 3.00 & 7.47 & 4.45 & 6.83 & 7.32 \\
\texttt{max\_independent\_set\_cpsat} & 5.17 & 3.41 & 3.61 & 3.21 & 1.90 \\
\texttt{max\_weighted\_independent\_set} & 1.25 & 1.20 & 1.72 & 1.23 & 1.13 \\
\texttt{min\_dominating\_set} & 1.09 & 1.91 & 1.23 & 1.75 & 1.52 \\
\texttt{min\_weight\_assignment} & 1.19 & 1.76 & 1.32 & 1.46 & 1.14 \\
\texttt{minimum\_spanning\_tree} & 4.16 & 6.52 & 3.39 & 4.32 & 1.84 \\
\texttt{minimum\_volume\_ellipsoid} & 3.71 & 20.67 & 20.30 & 12.31 & 3.06 \\
\texttt{multi\_dim\_knapsack} & 2.94 & 19.15 & 10.53 & 10.31 & 3.07 \\
\texttt{nmf} & 1.03 & 1.27 & 1.13 & 1.23 & 1.23 \\
\texttt{ode\_brusselator} & 7.08 & 4.08 & 3.00 & 3.71 & 1.97 \\
\texttt{ode\_fitzhughnagumo} & 1.05 & 1.14 & 1.10 & 1.09 & 1.04 \\
\texttt{ode\_hires} & 6.00 & 18.36 & 30.37 & 14.97 & 6.09 \\
\texttt{ode\_hodgkinhuxley} & 22.73 & 43.41 & 97.11 & 38.21 & 13.10 \\
\texttt{ode\_lorenz96\_nonchaotic} & 1.57 & 2.86 & 3.17 & 2.52 & 1.93 \\
\texttt{ode\_lotkavolterra} & 18.70 & 13.30 & 51.79 & 14.89 & 6.37 \\
\texttt{ode\_nbodyproblem} & 18.33 & 35.92 & 34.02 & 27.55 & 16.67 \\
\texttt{ode\_seirs} & 28.06 & 7.09 & 46.26 & 12.15 & 13.29 \\
\texttt{ode\_stiff\_robertson} & 4.66 & 23.90 & 6.52 & 12.89 & 4.31 \\
\texttt{ode\_stiff\_vanderpol} & 29.20 & 50.03 & 6.63 & 24.76 & 5.25 \\
\texttt{odr} & 1.01 & 1.02 & 1.01 & 1.02 & 1.01 \\
\texttt{optimal\_advertising} & 1.04 & 1.34 & 1.10 & 1.25 & 1.07 \\
\texttt{outer\_product} & 1.14 & 1.03 & 1.23 & 1.02 & 1.01 \\
\texttt{pagerank} & 1.50 & 2.05 & 18.29 & 3.34 & 4.00 \\
\texttt{pca} & 2.38 & 4.27 & 4.06 & 3.08 & 1.69 \\
\texttt{pde\_burgers1d} & 1.38 & 3.85 & 5.07 & 3.77 & 3.30 \\
\texttt{pde\_heat1d} & 1.32 & 2.01 & 1.44 & 1.76 & 1.25 \\
\texttt{polynomial\_mixed} & 5.30 & 5.84 & 3.23 & 4.27 & 2.16 \\
\texttt{polynomial\_real} & 7.38 & 39.65 & 4.55 & 16.70 & 3.33 \\
\texttt{power\_control} & 25.10 & 149.57 & 131.41 & 110.22 & 32.83 \\
\texttt{procrustes} & 1.38 & 1.30 & 1.30 & 1.27 & 1.14 \\
\texttt{psd\_cone\_projection} & 3.45 & 5.32 & 11.22 & 6.18 & 6.39 \\
\texttt{qp} & 1.46 & 1.98 & 2.79 & 2.02 & 1.74 \\
\texttt{qr\_factorization} & 2.10 & 1.72 & 1.84 & 1.66 & 1.45 \\
\texttt{quantile\_regression} & 1.18 & 1.40 & 2.01 & 1.41 & 1.29 \\
\texttt{queens\_with\_obstacles} & 1.77 & 3.31 & 4.31 & 3.20 & 2.11 \\
\texttt{queuing} & 1.07 & 1.04 & 1.08 & 1.05 & 1.02 \\
\texttt{qz\_factorization} & 1.23 & 1.19 & 1.20 & 1.18 & 1.09 \\
\texttt{randomized\_svd} & 2.27 & 3.80 & 2.28 & 2.91 & 1.52 \\
\texttt{rbf\_interpolation} & 1.01 & 1.01 & 1.01 & 1.01 & 1.00 \\
\texttt{rectanglepacking} & 1.40 & 1.39 & 3.02 & 1.48 & 1.27 \\
\texttt{robust\_kalman\_filter} & 2.59 & 3.87 & 9.57 & 4.39 & 3.66 \\
\texttt{robust\_linear\_program} & 2.32 & 1.96 & 7.62 & 2.40 & 1.82 \\
\texttt{rocket\_landing\_optimization} & 1.08 & 1.64 & 1.24 & 1.42 & 1.13 \\
\texttt{rotate\_2d} & 1.00 & 1.00 & 1.00 & 1.00 & 1.00 \\
\texttt{set\_cover} & 4.56 & 6.71 & 4.84 & 4.93 & 2.13 \\
\texttt{set\_cover\_conflicts} & 2.12 & 2.83 & 3.48 & 2.75 & 2.34 \\
\texttt{sha256\_hashing} & 1.00 & 1.00 & 1.00 & 1.00 & 1.00 \\
\texttt{shift\_2d} & 1.00 & 1.00 & 1.00 & 1.00 & 1.00 \\
\texttt{shortest\_path\_dijkstra} & 1.76 & 2.70 & 4.21 & 2.43 & 1.53 \\
\texttt{sinkhorn} & 1.43 & 2.09 & 3.29 & 1.98 & 1.42 \\
\texttt{sparse\_eigenvectors\_complex} & 1.00 & 1.00 & 1.00 & 1.00 & 1.00 \\
\texttt{sparse\_lowest\_eigenvalues\_posdef} & 1.57 & 2.33 & 2.78 & 2.19 & 1.87 \\
\texttt{sparse\_lowest\_eigenvectors\_posdef} & 1.59 & 2.76 & 3.06 & 2.44 & 1.79 \\
\texttt{sparse\_pca} & 2.95 & 7.61 & 8.47 & 5.75 & 2.91 \\
\texttt{spectral\_clustering} & 4.34 & 8.79 & 16.33 & 7.12 & 2.49 \\
\texttt{stable\_matching} & 1.45 & 1.89 & 2.55 & 1.91 & 1.64 \\
\texttt{svd} & 1.23 & 1.20 & 1.19 & 1.19 & 1.09 \\
\texttt{svm} & 1.00 & 1.00 & 1.00 & 1.00 & 1.00 \\
\texttt{sylvester\_solver} & 1.02 & 1.01 & 1.02 & 1.01 & 1.01 \\
\texttt{tensor\_completion\_3d} & 16.11 & 24.58 & 43.48 & 20.36 & 6.89 \\
\texttt{toeplitz\_solver} & 1.00 & 1.00 & 1.00 & 1.00 & 1.00 \\
\texttt{tsp} & 1.19 & 1.42 & 1.31 & 1.33 & 1.21 \\
\texttt{two\_eigenvalues\_around\_0} & 1.59 & 2.08 & 2.99 & 2.14 & 1.79 \\
\texttt{unit\_simplex\_projection} & 1.87 & 3.58 & 1.84 & 2.58 & 1.45 \\
\texttt{upfirdn1d} & 1.02 & 1.17 & 1.05 & 1.12 & 1.03 \\
\texttt{vector\_quantization} & 1.00 & 1.00 & 1.00 & 1.01 & 1.01 \\
\texttt{vectorized\_newton} & 1.00 & 1.00 & 1.00 & 1.00 & 1.00 \\
\texttt{vehicle\_routing} & 1.26 & 1.59 & 3.80 & 1.79 & 1.61 \\
\texttt{vertex\_cover} & 1.42 & 1.67 & 3.80 & 1.75 & 1.33 \\
\texttt{voronoi\_diagram} & 2.49 & 2.27 & 4.97 & 2.55 & 2.43 \\
\texttt{wasserstein\_dist} & 5.09 & 9.81 & 10.77 & 7.37 & 4.74 \\
\texttt{water\_filling} & 69.41 & 97.82 & 203.62 & 109.59 & 95.36 \\
\texttt{zoom\_2d} & 1.00 & 1.00 & 1.00 & 1.00 & 1.00 \\
\midrule
Harmonic mean & 1.60 & 1.89 & 1.94 & 1.84 & 1.59 \\
\bottomrule
\end{tabular}
\end{table*}

\subsection{ALEBench}
\label{sec:alebench_per_problem_performance}

Tables~\ref{tab:alebench_tasks_openevolve} and~\ref{tab:alebench_tasks_shinkaevolve} provide the ALE-Bench LITE per-problem results. GPT-5.4 achieves the strongest overall scores, followed by Qwen 3.7 Max and Gemini 3 Pro, while ShinkaEvolve is generally stronger across models and tasks. Performance varies considerably across problems and formats, with short contests typically scoring higher, although strong results on long-format tasks such as \texttt{ahc027} show that the method is not limited to shorter search problems.

\begin{table*}[t]
    \centering
    \footnotesize
    \setlength{\tabcolsep}{4pt}
    \caption{Per-task performance on ALE-Bench LITE using OpenEvolve+\sysname{}.}
    \label{tab:alebench_tasks_openevolve}
    \resizebox{0.8\textwidth}{!}{%
    \begin{tabular}{llccccc}
        \toprule
        Problem & Format
        & GPT-5.4
        & DeepSeek V4 Pro
        & Gemini 3 Pro
        & Qwen 3.7 Max
        & Qwen 3 Coder \\
        \midrule
        ahc008 & Long  & 1206.4 & 660.1  & 830.8  & 777.5  & 258.5 \\
        ahc011 & Long  & 1767.2 & 1177.8 & 1287.5 & 1388.6 & 584.1 \\
        ahc015 & Short & 1508.7 & 870.8  & 1154.4 & 1061.0 & 411.7 \\
        ahc016 & Long  & 1306.9 & 743.3  & 1055.4 & 847.9  & 295.2 \\
        ahc024 & Short & 1900.3 & 1313.3 & 1358.2 & 1491.4 & 715.2 \\
        ahc025 & Long  & 1495.7 & 885.2  & 1166.4 & 1032.3 & 369.6 \\
        ahc026 & Short & 1866.1 & 1123.4 & 1309.4 & 1432.3 & 602.9 \\
        ahc027 & Long  & 2172.8 & 1586.6 & 1580.9 & 1853.7 & 806.1 \\
        ahc039 & Short & 1989.1 & 1348.7 & 1421.7 & 1485.4 & 715.6 \\
        ahc046 & Short & 1885.8 & 1168.8 & 1431.3 & 1379.9 & 597.1 \\
        \bottomrule
    \end{tabular}
    }
\end{table*}

\begin{table*}[t]
    \centering
    \footnotesize
    \setlength{\tabcolsep}{4pt}
    \caption{Per-task performance on ALE-Bench LITE using ShinkaEvolve+\sysname{}.}
    \label{tab:alebench_tasks_shinkaevolve}
    \resizebox{0.8\textwidth}{!}{%
    \begin{tabular}{llccccc}
        \toprule
        Problem & Format
        & GPT-5.4
        & DeepSeek V4 Pro
        & Gemini 3 Pro
        & Qwen 3.7 Max
        & Qwen 3 Coder \\
        \midrule
        ahc008 & Long  & 1254.7 & 683.3  & 851.7  & 793.5  & 277.9 \\
        ahc011 & Long  & 1801.8 & 1219.1 & 1306.8 & 1445.6 & 609.8 \\
        ahc015 & Short & 1537.5 & 915.2  & 1169.1 & 1095.7 & 434.3 \\
        ahc016 & Long  & 1359.2 & 769.4  & 1071.1 & 865.4  & 314.3 \\
        ahc024 & Short & 1936.7 & 1353.2 & 1389.2 & 1525.0 & 739.7 \\
        ahc025 & Long  & 1525.1 & 934.7  & 1183.9 & 1053.5 & 385.8 \\
        ahc026 & Short & 1901.8 & 1157.6 & 1352.6 & 1464.5 & 623.5 \\
        ahc027 & Long  & 2171.2 & 1609.5 & 1604.5 & 1873.0 & 816.2 \\
        ahc039 & Short & 2027.1 & 1389.7 & 1439.7 & 1518.8 & 740.0 \\
        ahc046 & Short & 1921.9 & 1204.3 & 1449.4 & 1411.0 & 617.5 \\
        \bottomrule
    \end{tabular}
    }
\end{table*}

\section{Prompts}
\label{app:prompts}

The prompts used by \sysname{} are provided in Prompts~\ref{prompt:problem_insight}, \ref{prompt:change_insight}, \ref{prompt:cluster_summary}, and~\ref{prompt:recommendation_synthesis}. Prompt~\ref{prompt:problem_insight} summarizes the main bottleneck from the initial evaluation, while Prompt~\ref{prompt:change_insight} describes each modification and its observed fitness effect. Prompt~\ref{prompt:cluster_summary} abstracts the recurring optimization pattern shared by related insights. Finally, Prompt~\ref{prompt:recommendation_synthesis} converts retrieved insights, similarity scores, and linked evidence into exactly three grounded recommendations for the next mutation.

\begin{promptbox*}{Problem-Level Insight Generation}
\label{prompt:problem_insight}

You summarize performance bottlenecks at a high level from a single raw evolution record.

Write a localized insight that identifies the primary bottleneck and the main challenge reflected in the evaluation report.

\textbf{Requirements}
\begin{itemize}
    \item Write 2--4 concise sentences.
    \item Focus on algorithmic or architectural issues.
    \item Do not mention source code, file names, identifiers, or line-level details.
    \item Ground the insight only in the provided fitness score and evaluation report.
\end{itemize}

\textbf{Fitness score}

\{\{fitness\_score\}\}

\textbf{Raw evaluation report}

\{\{raw\_report\}\}

Return only the resulting insight without additional explanation.

\end{promptbox*}

\begin{promptbox*}{Change-Level Insight Generation}
\label{prompt:change_insight}

You summarize a localized implementation change from a single raw evolution record at a high level.

Write an insight describing the type of change and its observed impact on performance.

\textbf{Requirements}
\begin{itemize}
    \item Write 1--2 concise sentences.
    \item Describe the change using general concepts such as caching, pruning, vectorization, data-structure modification, or recursion reduction.
    \item Do not mention specific source-code identifiers, file names, or line-level details.
    \item Ground the stated impact in the parent fitness, child fitness, fitness delta, change summary, and evaluation report.
\end{itemize}

\textbf{Parent fitness}

\{\{parent\_fitness\}\}

\textbf{Child fitness}

\{\{child\_fitness\}\}

\textbf{Fitness delta}

\{\{fitness\_delta\}\}

\textbf{Changes summary}

\{\{changes\_summary\}\}

\textbf{Raw evaluation report}

\{\{raw\_report\}\}

Return only the resulting insight without additional explanation.

\end{promptbox*}

\begin{promptbox*}{Cluster Summary Generation}
\label{prompt:cluster_summary}

You summarize a cluster of semantically related local optimization insights.

Produce a concise practical summary that captures the shared optimization pattern, its recurring effect, and the general lesson represented by the cluster.

\textbf{Requirements}
\begin{itemize}
    \item Write 2--3 concise sentences.
    \item Generalize across the provided member insights rather than repeating them individually.
    \item Preserve important distinctions when the reported effects differ.
    \item Do not introduce claims that are unsupported by the member insights.
\end{itemize}

\textbf{Cluster member insights}

\{\{member\_insights\}\}

Return only the cluster summary without additional explanation.

\end{promptbox*}

\begin{promptbox*}{Recommendation Synthesis}
\label{prompt:recommendation_synthesis}

You synthesize actionable guidance from retrieved local insights and their associated raw evidence.

Given the current problem and retrieved evidence, produce concrete recommendations for the next mutation attempt.

\textbf{Requirements}
\begin{itemize}
    \item Return 3 concise and actionable recommendations.
    \item Ground every recommendation in the retrieved insights or linked raw evidence.
    \item Focus on changes that can guide the next code mutation.
    \item Avoid unsupported claims, speculative conclusions, and irrelevant advice.
    \item Do not merely repeat the retrieved insights; translate them into practical next steps.
\end{itemize}

\textbf{Current query or problem}

\{\{query\_text\}\}

\textbf{Retrieved evidence}

Each retrieved item may contain an insight, its cluster-level similarity, its insight-level similarity, and optional linked raw evidence.

\{\{retrieved\_evidence\}\}

Return only the numbered or bulleted recommendations without additional explanation.

\end{promptbox*}

\end{document}